\PassOptionsToPackage{table}{xcolor}
\documentclass{article}
\newif\ifreport
\reportfalse
\usepackage[T1]{fontenc}
\ifreport
  \newcommand{\mendrunninghead}{}
  \usepackage{mend_arxiv}
  \usepackage{lmodern}
\else
  \newcommand{\mendrunninghead}{Preprint}
  \newcommand{\mendtitlefont}{\fontsize{12.9pt}{16pt}\selectfont\sc}
  \usepackage{mend_arxiv,times}
\fi
\usepackage{microtype}

\usepackage{amsmath,amsfonts,bm}

\def\eqref#1{equation~\ref{#1}}

\def\1{\bm{1}}

\DeclareMathAlphabet{\mathsfit}{\encodingdefault}{\sfdefault}{m}{sl}
\SetMathAlphabet{\mathsfit}{bold}{\encodingdefault}{\sfdefault}{bx}{n}

\newcommand{\E}{\mathbb{E}}

\newcommand{\hpCapQ}{0.75}

\newcommand{\hpK}{3}
\newcommand{\hpEtaOne}{0.1}
\newcommand{\hpEtaTwo}{0.2}
\newcommand{\hpEtaThree}{0.4}
\newcommand{\hpEtas}{\ensuremath{\{\hpEtaOne, \hpEtaTwo, \hpEtaThree\}}}

\newcommand{\hpTauLoPct}{30\%}
\newcommand{\hpTauHiPct}{60\%}

\newcommand{\hpTq}{0.278}

\newcommand{\hpLambdaKeep}{10}
\newcommand{\hpMuRate}{0.001}
\newcommand{\hpMuMax}{0.5}

\newcommand{\hpLoraRank}{32}

\newcommand{\hpLR}{\ensuremath{3\times10^{-4}}}

\newcommand{\hpAdamEps}{\ensuremath{10^{-12}}}

\newcommand{\hpRolloutSteps}{10}

\newcommand{\hpRes}{512}

\newcommand{\hpPromptsO}{48}
\newcommand{\hpImagesO}{24}

\newcommand{\hpUpdatesO}{100}
\newcommand{\hpGPUsO}{3}
\newcommand{\hpPromptsE}{16}
\newcommand{\hpImagesE}{8}

\newcommand{\hpUpdatesE}{50}

\newcommand{\hpEvalPrompts}{200}
\newcommand{\hpEvalSeeds}{5}

\newcommand{\hpEvalSteps}{40}

\newcommand{\hpRefCFG}{4.5}

\input{results.tex}
\usepackage{amsmath,amssymb,amsthm,booktabs,graphicx,xcolor,multirow}
\usepackage{algorithm}
\usepackage{algpseudocode}
\usepackage[colorlinks=true,linkcolor=blue,citecolor=blue,urlcolor=blue]{hyperref}
\usepackage{url}
\usepackage{wrapfig,enumitem,needspace}
\makeatletter
\newcommand{\captionof}[1]{\def\@captype{#1}\caption}
\makeatother
\usepackage{etoolbox}
\ifreport
  \setcitestyle{numbers,square,comma}
  \usepackage[most]{tcolorbox}
  \definecolor{repabsbg}{HTML}{EFE7D8}
  \definecolor{repaccent}{HTML}{BF5700}
  \definecolor{reptitle}{HTML}{0F172A}
  \definecolor{replink}{HTML}{173A5E}
  \hypersetup{colorlinks=true,citecolor=repaccent,linkcolor=replink,urlcolor=repaccent}
  \newcommand{\heros}{\fontfamily{qhv}\selectfont}
  \makeatletter
  \def\section{\@startsection{section}{1}{\z@}{-2.0ex plus -0.5ex minus -.2ex}{1.5ex plus 0.3ex minus 0.2ex}{\large\bfseries\raggedright\color{repaccent}}}
  \def\subsection{\@startsection{subsection}{2}{\z@}{-1.8ex plus -0.5ex minus -.2ex}{0.8ex plus .2ex}{\normalsize\bfseries\raggedright\color{repaccent}}}
  \def\@maketitle{}
  \renewcommand{\headrulewidth}{0pt}
  \makeatother
  \newtcolorbox{titlepanel}{enhanced,breakable,colback=repabsbg,colframe=repaccent!35,boxrule=0.4pt,arc=2.5mm,
    left=12pt,right=12pt,top=10pt,bottom=10pt,boxsep=0pt,before skip=0pt,after skip=10pt}
\fi

\newtheorem{theorem}{Theorem}
\theoremstyle{definition}
\newtheorem{proposition}[theorem]{Proposition}
\theoremstyle{plain}

\theoremstyle{definition}

\theoremstyle{remark}

\newcommand{\mend}{\textup{\textsc{MEND}}}
\definecolor{mendblue}{RGB}{30,90,170}
\definecolor{goodblue}{RGB}{30,90,170}
\definecolor{badred}{RGB}{200,60,60}
\newcommand{\tbd}{\ph{\textsc{tbd}}}
\definecolor{phred}{RGB}{139,0,0}
\newcommand{\ph}[1]{{\setlength{\fboxsep}{0.6pt}\setlength{\fboxrule}{0.4pt}\fcolorbox{phred}{white}{\textcolor{phred}{#1}}}}
\definecolor{tabhead}{RGB}{232,236,243}
\definecolor{tabours}{RGB}{232,240,250}
\definecolor{tabgroup}{RGB}{246,246,246}
\newcommand{\oursrow}{\rowcolor{tabours}}
\newcommand{\headrow}{\rowcolor{tabhead}}

\newcommand{\best}[1]{\textbf{#1}}
\newcommand{\second}[1]{\underline{#1}}
\newcommand{\placeholderfig}[3]{\fcolorbox{red}{white}{\parbox[c][\dimexpr#2-2\fboxsep-2\fboxrule][c]{\dimexpr#1-2\fboxsep-2\fboxrule}{\centering\footnotesize\textcolor{red}{\textsc{placeholder}: #3}}}}
\newcommand{\mendfig}[4]{\IfFileExists{figures/#1}{\includegraphics[width=#2]{figures/#1}}{\placeholderfig{#2}{#3}{\texttt{\detokenize{figures/#1}}\\[2pt] #4}}}

\newcommand{\Rk}{R_\kappa}
\newcommand{\norm}[1]{\lVert #1 \rVert}

\newcommand{\sg}{\operatorname{sg}}

\newcommand{\vold}{v_{\mathrm{old}}}
\newcommand{\xhat}{\hat{x}}

\newcommand{\ystar}{y^{\star}}

\newcommand{\papertitle}{\mend{}: RL for Flow Models via Proximal Velocity Matching}
\title{\papertitle}

\author{Shreshth Saini$^{1,2}$ \quad Neil Birkbeck$^{2}$ \quad Yilin Wang$^{2}$ \quad Balu Adsumilli$^{2}$ \quad Alan C.\ Bovik$^{1,3}$\\[3pt]
{\normalfont $^{1}$The University of Texas at Austin \qquad $^{2}$Google \qquad $^{3}$University of Colorado Boulder}\\[1pt]
{\normalfont\texttt{saini.2@utexas.edu}}}

\begin{document}
\addtocontents{toc}{\protect\setcounter{tocdepth}{-1}}
\raggedbottom
\maketitle

\ifreport
\vspace*{-0.35in}
\begin{titlepanel}
{\heros\bfseries\fontsize{15.5pt}{19pt}\selectfont\raggedright\color{reptitle}MEND: RL for Flow Models via Proximal Velocity Matching\par}
\vskip 8pt
{\raggedright
{\heros\bfseries\normalsize Shreshth Saini\textsuperscript{1,2}\quad Neil Birkbeck\textsuperscript{2}\quad Yilin Wang\textsuperscript{2}\quad Balu Adsumilli\textsuperscript{2}\quad Alan C.\ Bovik\textsuperscript{1,3}\par}
\vskip 3pt
{\small \textsuperscript{1}The University of Texas at Austin \qquad \textsuperscript{2}Google \qquad \textsuperscript{3}University of Colorado Boulder\par}
\par}
\vskip 9pt
{\small Reward post-training of flow models either reweights the model's own samples under a KL penalty or a frozen reference, often for thousands of updates, or backpropagates the reward and moves every sample without checking that the move is worth its size. We introduce \mend{}, a reinforcement learning method built on proximal velocity matching. \mend{} caps rewards within each prompt group, so samples that already score well receive no move. Below the cap, it proposes moves along the reward gradient and accepts one only when its capped reward gain exceeds a quadratic displacement price. The model then regresses onto the resulting velocity targets, with no KL term, frozen reference model, or advantage weights. In \hpUpdatesO{} updates, \mend{} outperforms Flow-GRPO (about \res{flow.updates} updates) on five of six evaluators at the same distance to base-model images. Under an equal-budget protocol, it surpasses ReFL and DiffusionNFT at every evaluated update across four training rewards, reaching PickScore \mPick{} versus \res{refl.pick} and \res{nftequal.pick}, respectively. A 300-update three-reward run also surpasses the five-reward DiffusionNFT model on all three rewards it trains on. \mend{} is general and easy to adopt: it applies to any flow backbone with a differentiable reward. \href{https://shreshthsaini.github.io/MEND-RL/}{\textbf{Code}}
\par}
\vskip 9pt
{\small
\noindent
\begin{minipage}[c]{0.63\linewidth}
\textbf{Date:} \today\\[2.5pt]
\textbf{Correspondence:} \href{mailto:saini.2@utexas.edu}{\textcolor{repaccent}{\texttt{saini.2@utexas.edu}}}\\[2.5pt]
\textbf{Code:} \href{https://github.com/shreshthsaini/MEND-RL}{\textcolor{repaccent}{\texttt{github.com/shreshthsaini/MEND-RL}}}
\end{minipage}\hfill
\begin{minipage}[c]{0.35\linewidth}
\raggedleft
\raisebox{-0.5\height}{\includegraphics[height=12pt]{assets/youtube.png}}%
\hspace{10pt}%
\raisebox{-0.5\height}{\includegraphics[height=16pt]{assets/utaustin.png}}
\end{minipage}}
\end{titlepanel}
\else
\begin{abstract}
Reward post-training of flow models either reweights the model's own samples under a KL penalty or a frozen reference, often for thousands of updates, or backpropagates the reward and moves every sample without checking that the move is worth its size. We introduce \mend{}, a reinforcement learning method built on proximal velocity matching. \mend{} caps rewards within each prompt group, so samples that already score well receive no move. Below the cap, it proposes moves along the reward gradient and accepts one only when its capped reward gain exceeds a quadratic displacement price. The model then regresses onto the resulting velocity targets, with no KL term, frozen reference model, or advantage weights. In \hpUpdatesO{} updates, \mend{} outperforms Flow-GRPO (about \res{flow.updates} updates) on five of six evaluators at the same distance to base-model images. Under an equal-budget protocol, it surpasses ReFL and DiffusionNFT at every evaluated update across four training rewards, reaching PickScore \mPick{} versus \res{refl.pick} and \res{nftequal.pick}, respectively. A 300-update three-reward run also surpasses the five-reward DiffusionNFT model on all three rewards it trains on. \mend{} is general and easy to adopt: it applies to any flow backbone with a differentiable reward. \href{https://shreshthsaini.github.io/MEND-RL/}{\textbf{Code}}

\end{abstract}
\fi
\begin{figure}[H]
\centering
\begin{minipage}{0.95\linewidth}
\centering
\mendfig{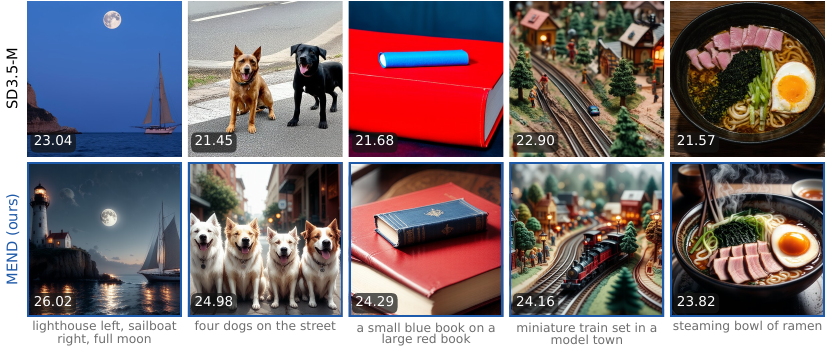}{\linewidth}{2.36in}{Selected base and \mend{} pairs}\par\smallskip
\mendfig{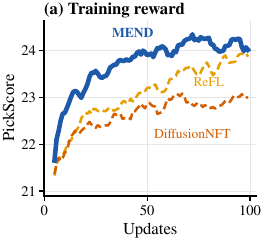}{0.32\linewidth}{1.6in}{(a) Training reward}\hfill
\mendfig{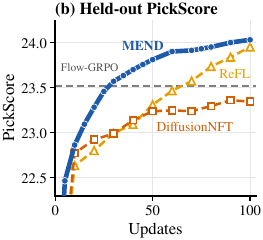}{0.32\linewidth}{1.6in}{(b) Held-out PickScore}\hfill
\mendfig{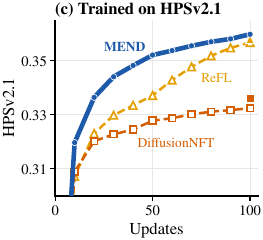}{0.32\linewidth}{1.6in}{(c) Trained on HPSv2.1}
\end{minipage}
\caption{\textbf{\mend{} stays ahead of ReFL and DiffusionNFT at every evaluated update.} Top: SD3.5-M above, \mend{} below, same prompt and initial noise. Bottom, equal budget: (a) training reward and (b) held-out PickScore against ReFL and DiffusionNFT under the same protocol, with Flow-GRPO (about \res{flow.updates} updates) at its own setting as a level; (c) the HPSv2.1 run's trained reward.}
\label{fig:hero}
\end{figure}

\section{Introduction}\label{sec:intro}
Learned reward models make it possible to post-train text-to-image flow models for human preference \citep{kirstain2023pickapic,wu2023hpsv2,xu2023imagereward}. The open problem is how to turn a scalar score into an update that raises reward quickly without eroding what the base model already does well. A score alone does not say which samples should change or where they should go. Policy-gradient methods such as Flow-GRPO reweight stochastic trajectories by group-relative advantages under a KL penalty \citep{liu2025flowgrpo}. Weighted-regression methods such as DiffusionNFT reweight the model's own samples inside a regression loss \citep{zheng2025nft}. Both adjust how much existing samples count, and the Flow-GRPO and DiffusionNFT models we compare against use about \res{flow.updates} and \res{nft.updates} updates.

\begin{figure}[t]
\centering
\includegraphics[width=0.96\linewidth]{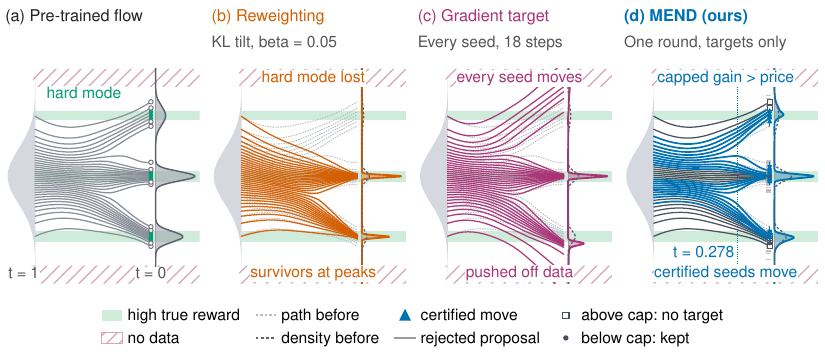}
\caption{(a) The base flow. (b) Reward reweighting concentrates on the peaks and loses the hard mode. (c) Gradient ascent moves every sample and pushes some of the mass off the data. (d) \mend{}: capped samples stay, certified moves are short, and the hard mode keeps its mass (Appendix~\ref{app:hero-toy}).}
\label{fig:toy}
\end{figure}

\begin{wraptable}{r}{0.5\linewidth}
\vspace{-9pt}
\caption{\textbf{Objective components.} Blue matches \mend{}; red differs. The check compares reward gain with displacement cost. \mend{} uses a lagged behavior adapter, not a frozen reference. Updates refer to the compared runs.}
\label{tab:paradigm}
\small
\setlength{\tabcolsep}{3pt}
\resizebox{\linewidth}{!}{%
\begin{tabular}{@{}lccccc@{}}
\toprule
Method & \shortstack{KL / frozen\\reference} & \shortstack{Reward\\weights} & $\nabla R$ & \shortstack{Target\\check} & Updates \\
\midrule
Flow-GRPO & \textcolor{badred}{yes} & \textcolor{badred}{yes} & \textcolor{badred}{no} & \textcolor{badred}{no} & $\sim$\res{flow.updates} \\
ReFL & \textcolor{goodblue}{no} & \textcolor{goodblue}{no} & \textcolor{goodblue}{yes} & \textcolor{badred}{no} & \hpUpdatesO \\
DiffusionNFT & \textcolor{badred}{yes} & \textcolor{badred}{yes} & \textcolor{badred}{no} & \textcolor{badred}{no} & \res{nft.updates} \\
\oursrow \mend{} & \textcolor{goodblue}{no} & \textcolor{goodblue}{no} & \textcolor{goodblue}{yes} & \textcolor{goodblue}{yes} & \hpUpdatesO \\
\bottomrule
\end{tabular}}
\vspace{-6pt}
\end{wraptable}
Reward backpropagation supplies a direction: ReFL and DRaFT differentiate the reward through sampling \citep{xu2023imagereward,clark2024draft}. But the gradient moves every sample, including those that already score well, and nothing checks that a move earns its size. A KL penalty or reference term limits drift of the distribution as a whole and makes no per-sample decision. The three-mode example of Figure~\ref{fig:toy} shows both failures: reweighting suppresses a hard mode, and unchecked ascent pushes samples into low-density regions.

\begin{figure}[t]
\centering
\mendfig{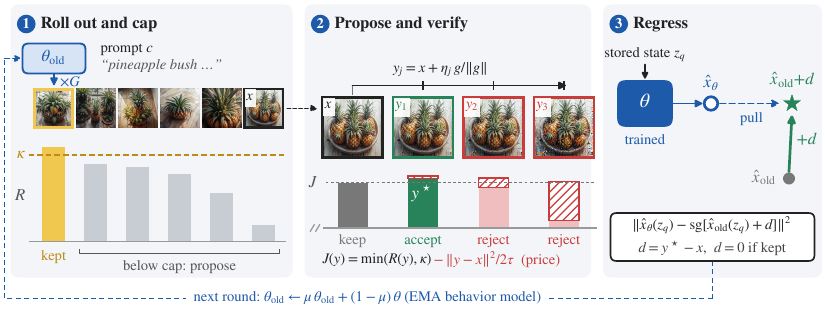}{\linewidth}{2.12in}{Roll out and cap; propose and verify; regress onto the accepted displacement}
\caption{\mend{} overview. (1) A group for prompt $c$; samples at or above the cap $\kappa$ are kept. (2) Each sample below the cap gets \hpK{} proposals along its reward gradient; the verdict subtracts a distance price from the capped reward and keeps the best candidate, here the shortest move $y_1$. (3) The trained adapter regresses at the stored state onto the behavior prediction plus the accepted displacement $d$, a velocity target by \eqref{eq:velocity}; the behavior model then follows it by EMA.}
\label{fig:method}
\end{figure}

We introduce \mend{}, an RL method for flow models based on \emph{proximal velocity matching} (Figure~\ref{fig:method}). Its principle is that a sample should move only when the reward it gains pays for the distance it travels. Within each prompt group, \mend{} caps rewards at a quantile, so samples that already score well keep their place. For each sample below the cap, it proposes a few moves along the normalized reward gradient and selects the candidate with the highest capped reward minus a quadratic displacement price, with the unchanged sample always among the candidates. This is a proximal-point step evaluated on decoded candidates. The model then regresses its velocity at a stored rollout state onto the behavior velocity shifted by the accepted displacement, with no KL term, frozen reference model, or advantage weights (Table~\ref{tab:paradigm}).
The selection rule comes with guarantees. Every accepted target improves capped reward by more than its quadratic price, and its displacement bound shrinks to zero as the sample approaches the cap (Proposition~\ref{prop:target}). At the behavior parameters, the descent direction of the regression loss backpropagates normalized reward gradients through the clean predictions of accepted samples only, scaled by their selected steps (Proposition~\ref{prop:firstorder}).

\begin{figure}[t]
\centering
\mendfig{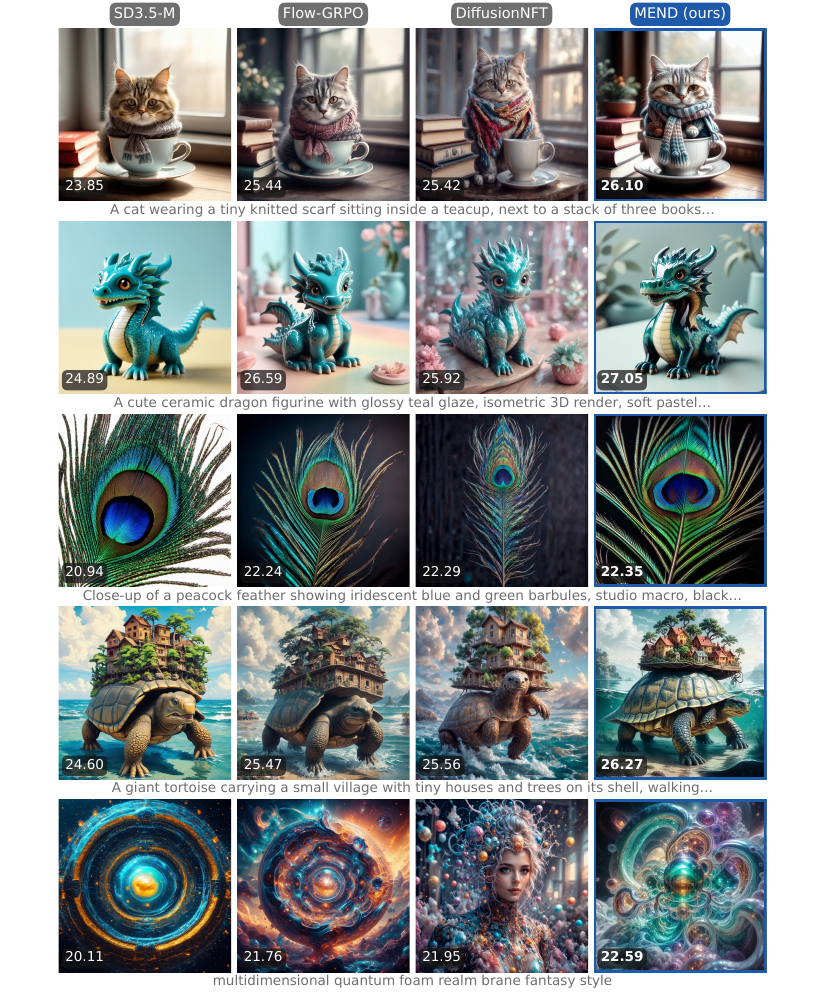}{0.76\linewidth}{0.62\textheight}{Matched prompts across SD3.5-M, Flow-GRPO, DiffusionNFT, and \mend{}}
\caption{Rows share a prompt and seed across SD3.5-M, Flow-GRPO, DiffusionNFT, and \mend{}. PickScore on each tile.}
\label{fig:cmp}
\end{figure}

After \hpUpdatesO{} PickScore updates, \mend{} outperforms Flow-GRPO, trained for about \res{flow.updates} updates, on every evaluator except aesthetic score, at the same distance to base-model images (Table~\ref{tab:main}). Under the equal-budget protocol of \citet{zhou2026opsd}, it surpasses ReFL and DiffusionNFT at every evaluated update on all four training rewards (Figures~\ref{fig:hero} and~\ref{fig:rewards}). Figures~\ref{fig:hero} and~\ref{fig:cmp} show matched visual comparisons. Our contributions are:

\begin{itemize}[leftmargin=*,nosep,topsep=3pt]
\item \textbf{Insight and method.} We observe that existing reward post-training never decides, sample by sample, whether a change is worth making. We propose \mend{}, which makes this decision explicit through proximal velocity matching: a sample receives a new velocity target only when its capped reward gain exceeds a quadratic displacement price, and the model is trained by plain regression onto these targets, with no KL term, frozen reference model, or advantage weights.
\item \textbf{Theory.} We prove that every accepted target improves reward by more than its displacement price, with a displacement bound that vanishes as the sample approaches the cap (Proposition~\ref{prop:target}). We further show that the resulting regression gradient is reward backpropagation restricted to the accepted samples and scaled by their selected steps (Proposition~\ref{prop:firstorder}).
\item \textbf{Performance and generality.} In only \hpUpdatesO{} updates, \mend{} outperforms Flow-GRPO (about \res{flow.updates} updates) on five of six evaluators at the same base distance. At equal budget, it surpasses ReFL and DiffusionNFT on all four training rewards, reaching PickScore \mPick{} versus \res{refl.pick} and \res{nftequal.pick}. \mend{} is backbone-agnostic and easy to adopt: the same method improves SD3.5-M, SD3-M and the distilled Z-Image-Turbo at the same update budget.
\end{itemize}

\section{Method}\label{sec:method}

\mend{} rests on one principle: a sample's target moves only when the reward it gains pays for the distance it travels. Each round applies this principle sample by sample to build targets, then fits them with a plain regression loss. A round consists of a rollout, target selection (cap, propose, verify), and one regression update (Algorithm~\ref{alg:mend}).

\begin{algorithm}[b]
\caption{\mend{}: one training round.}\label{alg:mend}
\small
\begin{algorithmic}[1]
\Require adapters $\theta$, $\theta_{\mathrm{old}}$; penalty scale $\tau$; cap quantile $q$; steps $\eta_1,\ldots,\eta_K$
\State Roll out $\theta_{\mathrm{old}}$ for every prompt and seed; store endpoint $x$ and state $z_q$ \Comment{rollout}
\State Compute $R(x,c)$ and $g$ for all samples; set $\kappa(c)$ by \eqref{eq:cap} \Comment{cap}
\For{each sample with $R(x,c)<\kappa(c)$ and $g\ne0$}
  \State Form proposals by \eqref{eq:proposal}; select $\ystar$ by \eqref{eq:verdict}; $d\gets\ystar-x$ \Comment{propose, verify}
\EndFor
\State Set $d\gets0$ for all other samples; take one AdamW step on \eqref{eq:loss} \Comment{regress}
\State Update $\tau$ from the acceptance rate; $\theta_{\mathrm{old}}\gets\mu_u\theta_{\mathrm{old}}+(1-\mu_u)\theta$ \Comment{penalty, behavior}
\end{algorithmic}
\end{algorithm}

\textbf{Setting and notation.} A rectified flow uses the interpolation $z_t=(1-t)x_0+t\epsilon$ between a clean latent $x_0$ and Gaussian noise $\epsilon$. For prompt $c$, its velocity $v_\theta(z_t,t,c)$ gives the clean prediction $\xhat_\theta(z_t)=z_t-t\,v_\theta(z_t,t,c)$ \citep{liu2022rectified,lipman2022flow,esser2024sd3}. Two adapters share the base model: the trained adapter $\theta$ and the behavior adapter $\theta_{\mathrm{old}}$. For each prompt, $\theta_{\mathrm{old}}$ generates $G$ deterministic \hpRolloutSteps{}-step trajectories \citep{lu2022dpmsolverpp}, with endpoints $x^{(1)},\ldots,x^{(G)}$ and stored states $z_q$ nearest $t_q=\hpTq$. We use the root-mean-square norm $\norm{\cdot}$ and its mean inner product $\langle\cdot,\cdot\rangle$. $R(x,c)$ scores the decoded latent, and $g=\nabla_x R(x,c)$ is its gradient under this inner product.

\textbf{Cap.} Reward chasing starts with samples that already score well: pushing them further buys little reward and costs diversity. Within each prompt group, we therefore cap rewards at the quantile $q=\hpCapQ$, subject to a global floor $\kappa_{\mathrm{glob}}$ that rises slowly across rounds (Appendix~\ref{app:settings}):
\begin{equation}\label{eq:cap}
\kappa(c)=\max\!\Big\{Q_q\big(\{R(x^{(k)},c)\}_{k=1}^G\big),\;\kappa_{\mathrm{glob}}\Big\},
\qquad \Rk(x,c)=\min\{R(x,c),\kappa(c)\}.
\end{equation}
Samples at or above the cap receive zero displacement, and $\Rk$ gives no credit for exceeding it.

\textbf{Propose.} Below the cap, the reward gradient gives a direction of improvement, and normalizing it makes each step $\eta_j$ a displacement length independent of the reward's scale. Each sample below the cap with $g\ne0$ receives $K=\hpK$ proposals along its normalized reward gradient, each decoded and scored once:
\begin{equation}\label{eq:proposal}
y_j=x+\eta_j\,\frac{g}{\norm{g}},\qquad \eta_j\in\hpEtas.
\end{equation}
Samples with $g=0$ are kept. The decision below thus rests on the reward attained at $y_j$, not on a first-order prediction of it.

\textbf{Verify.} A proposal must pay for its length. We select the candidate with the highest capped reward minus a quadratic displacement penalty, always including the unchanged sample:
\begin{equation}\label{eq:verdict}
\ystar=\arg\max_{y\in\{x,y_1,\ldots,y_K\}}J(y),
\qquad J(y)=\Rk(y,c)-\frac{\norm{y-x}^2}{2\tau}.
\end{equation}
This is the proximal-point objective \citep{rockafellar1976,jko1998}, maximized over a finite candidate set instead of all of latent space: $\tau$ sets the price of distance, and a move is accepted only if its capped gain exceeds $\norm{y-x}^2/(2\tau)$. Ties favor $x$, then the shortest proposal. The parameter $\tau>0$ is fixed within a round and adjusted between rounds to target an acceptance rate of \hpTauLoPct{} to \hpTauHiPct{} among proposed samples (Appendix~\ref{app:settings}).

\textbf{Regress.} Let $d=\ystar-x$ for accepted samples and $d=0$ otherwise. At the stored state $z_q$, we match the behavior prediction plus $d$, not the endpoint $\ystar$ itself:
\begin{equation}\label{eq:loss}
\mathcal L(\theta)=\E_{\mathrm{moved}}\norm{\xhat_\theta(z_q)-\sg[\xhat_{\mathrm{old}}(z_q)+d]}^2
+\lambda_{\mathrm{keep}}\,\E_{\mathrm{kept}}\norm{\xhat_\theta(z_q)-\sg[\xhat_{\mathrm{old}}(z_q)]}^2 .
\end{equation}
Here $\xhat_{\mathrm{old}}=\xhat_{\theta_{\mathrm{old}}}$, $\sg$ stops gradients, and $\lambda_{\mathrm{keep}}=\hpLambdaKeep$. Each expectation averages its subset of the round; an empty subset contributes zero. The endpoint $x$ differs from the one-step prediction $\xhat_{\mathrm{old}}(z_q)$ by the rest of the rollout, and regressing on $\ystar$ would mix that reward-independent gap into the update. With our target, a kept sample asks for no change and a moved sample asks for exactly the correction $d$.\looseness=-1

Since $\xhat_\theta(z_q)=z_q-t_q\,v_\theta(z_q,t_q,c)$, each term of \eqref{eq:loss} is a velocity-matching loss:
\begin{equation}\label{eq:velocity}
\norm{\xhat_\theta(z_q)-\sg[\xhat_{\mathrm{old}}(z_q)+d]}^2
=t_q^2\,\norm{v_\theta(z_q,t_q,c)-v^\star}^2,
\qquad v^\star=\sg\!\big[\vold(z_q,t_q,c)-d/t_q\big],
\end{equation}
with $\vold=v_{\theta_{\mathrm{old}}}$. Kept samples have $v^\star=\vold$. \mend{} is therefore velocity matching onto a target selected by the proximal rule \eqref{eq:verdict}, which is why we call it proximal velocity matching. We call $v^\star$, equivalently $\xhat_{\mathrm{old}}(z_q)+d$, the target.

\textbf{Behavior model and optimizer.} The behavior adapter follows $\theta$ with EMA coefficient $\mu_u=\min(\hpMuRate\,u,\hpMuMax)$ after update $u$ (Algorithm~\ref{alg:mend}, line~7). The keep term therefore anchors velocities to a lagged adapter that follows training, not to a frozen reference. Each round takes one AdamW step with constant learning rate \hpLR{} and $\epsilon=\hpAdamEps$; evaluation uses a separate EMA of $\theta$ (Algorithm~\ref{alg:impl}). Per update, \mend{} needs one reward backward pass per sample, at most \hpK{} decodes and reward evaluations per sample below the cap, and one network forward and backward pass per sample at $z_q$. It never backpropagates through the sampler.

\section{Theoretical analysis}\label{sec:theory}
Target selection evaluates a proximal objective \citep{rockafellar1976,jko1998} over a finite candidate set containing the unchanged sample. Two results follow: accepted moves are bounded by the reward they gain, and the regression update is reward backpropagation restricted to accepted samples. Proofs are in Appendix~\ref{app:proofs}.

\begin{proposition}[Verified target selection]\label{prop:target}
Fix a round's samples $x$, caps $\kappa$, price $\tau>0$, and proposal steps $0<\eta_1\le\cdots\le\eta_K$. Include $x$ among the candidates and accept a move only if its objective strictly exceeds that of $x$. Every accepted target satisfies
\begin{equation}\label{eq:certificate}
\frac{\norm{\ystar-x}^2}{2\tau}
\leq \Rk(\ystar)-\Rk(x)
\leq \kappa-\Rk(x).
\end{equation}
Hence $\norm{\ystar-x}\le\min\{\eta_K,\sqrt{2\tau(\kappa-\Rk(x))}\}$ and $R(\ystar)-R(x)\ge\norm{\ystar-x}^2/(2\tau)$.
\end{proposition}
The bound tightens as a sample nears the cap: its headroom $\kappa-\Rk(x)$ goes to zero, and so does the largest move it can receive.

\begin{proposition}[First-order realization]\label{prop:firstorder}
Fix a round's targets and rollout states, and let $\xhat_\theta$ be differentiable in $\theta$. Let $\mathcal A\ne\varnothing$ be the accepted seeds, with steps $\eta_i$, normalized endpoint reward gradients $\hat h_i=g_i/\norm{g_i}$, and stored states $z_q^{(i)}$. Then
\begin{equation}\label{eq:firstorder}
-\nabla\mathcal L(\theta_{\mathrm{old}})
=\frac{2}{|\mathcal A|}\sum_{i\in\mathcal A}\eta_i\,
\left.\nabla_\theta\big\langle \hat h_i,\,\xhat_\theta(z_q^{(i)})\big\rangle\right|_{\theta=\theta_{\mathrm{old}}},
\end{equation}
so kept seeds contribute zero gradient at $\theta_{\mathrm{old}}$.
\end{proposition}
Equation~\ref{eq:firstorder} backpropagates normalized endpoint reward gradients through the clean predictions, restricted to accepted seeds and scaled by their selected steps. \mend{} thus keeps the direction of reward backpropagation and adds the two decisions it lacks: which samples move, and how far.

\begin{table}[t]
\centering
\caption{\mend{} outperforms Flow-GRPO on five of six evaluators at the same base distance. Its three-reward run surpasses DiffusionNFT on those three rewards. DrawBench $200\times5$.}
\label{tab:main}
\small
\setlength{\tabcolsep}{3pt}
\resizebox{\linewidth}{!}{%
\begin{tabular}{@{}llcccccccc@{}}
\toprule
\headrow Method & Rewards & Updates & PickScore & HPSv2.1 & HPSv3 & \shortstack{Image\\Reward} & CLIPScore & Aesthetic & \shortstack{Dist. to\\base $\downarrow$} \\
\midrule
\textcolor{gray}{SD3.5-M} & \textcolor{gray}{none} & \textcolor{gray}{\res{base.updates}} & \textcolor{gray}{\res{base.pick}} & \textcolor{gray}{\res{base.hps2}} & \textcolor{gray}{\res{base.hps3}} & \textcolor{gray}{\res{base.ir}} & \textcolor{gray}{\res{base.clip}} & \textcolor{gray}{\res{base.aes}} & \textcolor{gray}{\res{base.dist}} \\
Flow-GRPO & PickScore & $\sim$\res{flow.updates} & \res{flow.pick} & \res{flow.hps2} & \res{flow.hps3} & \res{flow.ir} & \res{flow.clip} & \res{flow.aes} & \best{\res{flow.dist}} \\
DiffusionNFT & five & \res{nft.updates} & \second{\res{nft.pick}} & \second{\res{nft.hps2}} & \best{\res{nft.hps3}} & \best{\res{nft.ir}} & \second{\res{nft.clip}} & \best{\res{nft.aes}} & \res{nft.dist} \\
\oursrow \textbf{\mend{} (ours)} & PickScore & \hpUpdatesO & \mgPick & \res{cfg45.hps2} & \second{\res{cfg45.hps3}} & \second{\res{cfg45.ir}} & \mgClip & \res{cfg45.aes} & \best{\mgDist} \\
\oursrow \textbf{\mend{} (ours)} & three & 300 & \best{\res{o3.pick}} & \best{\res{o3.hps2}} & \res{o3.hps3} & \res{o3.ir} & \best{\res{o3.clip}} & \second{\res{o3.aes}} & \second{\res{o3.dist}} \\
\bottomrule
\end{tabular}}
\end{table}
\begin{figure}[t]
\centering
\mendfig{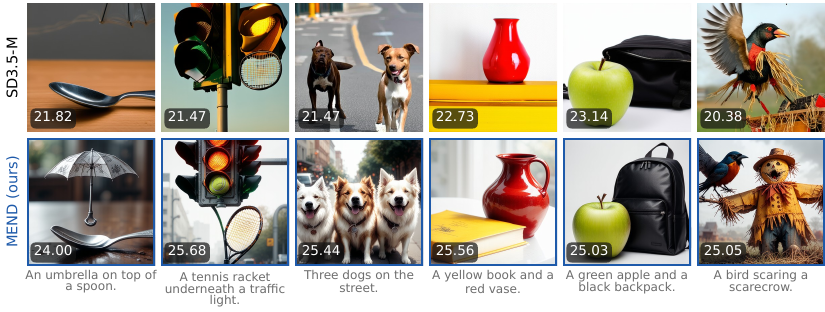}{0.92\linewidth}{1.9in}{DrawBench prompts on counts, attributes and relations}
\caption{DrawBench prompts (text under each column), SD3.5-M above and \mend{} below from the same initial noise, PickScore on each tile.}
\label{fig:fidelity}
\end{figure}

\section{Experiments}\label{sec:exp}
\subsection{Setup}\label{sec:setup}
We train SD3.5-M \citep{esser2024sd3} on PickScore under the equal-budget protocol of \citet{zhou2026opsd}: \hpPromptsO{} Pick-a-Pic prompts \citep{kirstain2023pickapic} $\times$ \hpImagesO{} images per update, rank-\hpLoraRank{} LoRA, \hpRolloutSteps-step rollouts at \hpRes{} pixels, and \hpUpdatesO{} updates. ReFL and DiffusionNFT are the baselines under this protocol. We also compare with the Flow-GRPO PickScore adapter \citep{liu2025flowgrpo} (about \res{flow.updates} updates) and the five-reward DiffusionNFT model \citep{zheng2025nft} (\res{nft.updates} updates). Evaluation uses DrawBench \citep{saharia2022imagen}, \hpEvalPrompts{} prompts $\times$ \hpEvalSeeds{} seeds and \hpEvalSteps{} Euler steps, scored with PickScore, HPSv2.1 \citep{wu2023hpsv2}, HPSv3 \citep{ma2025hpsv3}, ImageReward \citep{xu2023imagereward}, CLIPScore \citep{hessel2021clipscore} and aesthetic quality \citep{schuhmann2022laion}. Fidelity is the DreamSim distance \citep{fu2023dreamsim} to the base image generated with the same prompt, noise and sampling settings. Appendix~\ref{app:baselines} gives baseline settings, and Appendix~\ref{app:setup} repeats the main comparison on DrawBench prompts held out from all design decisions.

\subsection{Main comparison}\label{sec:main}
\mend{} outperforms Flow-GRPO on five of six evaluators at the same base distance (\mgDist{}), with roughly \res{update.ratio}-fold fewer updates (Table~\ref{tab:main}). Aesthetic score is the one evaluator on which Flow-GRPO is higher (\res{flow.aes} versus \res{cfg45.aes}). The run trained on PickScore, HPSv2.1 and CLIPScore surpasses DiffusionNFT (\res{nft.updates} updates on five rewards) on those three evaluators at a smaller base distance, and is lower on HPSv3, ImageReward and aesthetic score, which it does not train on. Table~\ref{tab:main} and Figures~\ref{fig:fidelity} to~\ref{fig:progress}, except Figure~\ref{fig:dynamics}(a), use each model's headline sampling setting; Table~\ref{tab:transfer} and Figures~\ref{fig:hero} and~\ref{fig:rewards} use the equal-budget protocol's setting (Appendix~\ref{app:settings}). Section~\ref{sec:frontier} compares equal budgets. Figure~\ref{fig:fidelity} shows DrawBench examples where \mend{} renders counts, attributes or spatial relations that SD3.5-M misses from the same noise.

\subsection{Update efficiency and fidelity}\label{sec:frontier}
\begin{figure}[t]
\centering
\mendfig{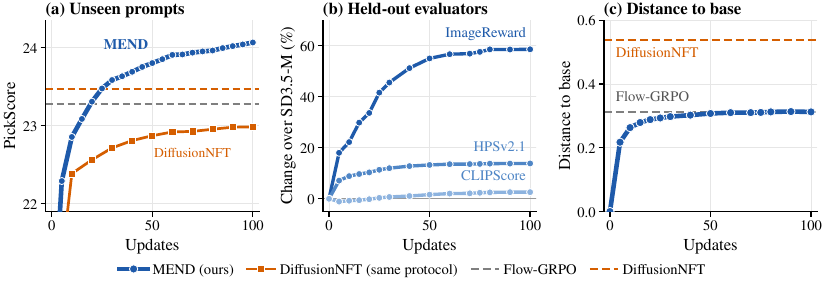}{\linewidth}{1.9in}{Unseen-prompt reward, held-out evaluators, and distance to the base}
\caption{(a) PickScore on 64 held-out Pick-a-Pic prompts at the protocol's setting, against DiffusionNFT under the same protocol, with Flow-GRPO and DiffusionNFT (about \res{flow.updates} and \res{nft.updates} updates) as levels; (b) held-out evaluators and (c) distance to the base on DrawBench at guidance \hpRefCFG{}.}
\label{fig:dynamics}
\end{figure}

At equal budget, \mend{} is ahead of both baselines at every evaluated update. In Figure~\ref{fig:hero}(b), it reaches PickScore \mPick{} at update \hpUpdatesO{}, versus \res{refl.pick} for ReFL and \res{nftequal.pick} for DiffusionNFT. On 64 unseen Pick-a-Pic prompts (Figure~\ref{fig:dynamics}(a)), it is above DiffusionNFT under the same protocol at every evaluated update. By update 25 it also surpasses the levels of Flow-GRPO and DiffusionNFT, which use about \res{flow.updates} and \res{nft.updates} updates, respectively. Training costs \res{train.gpuh} GPU-hours on \hpGPUsO{} GB200 GPUs, excluding evaluation, and the result is stable across launches: a second launch of the same configuration differs by 0.01 PickScore at update 50 (Appendix~\ref{app:sensitivity}).

\begin{figure}[t]
\centering
\mendfig{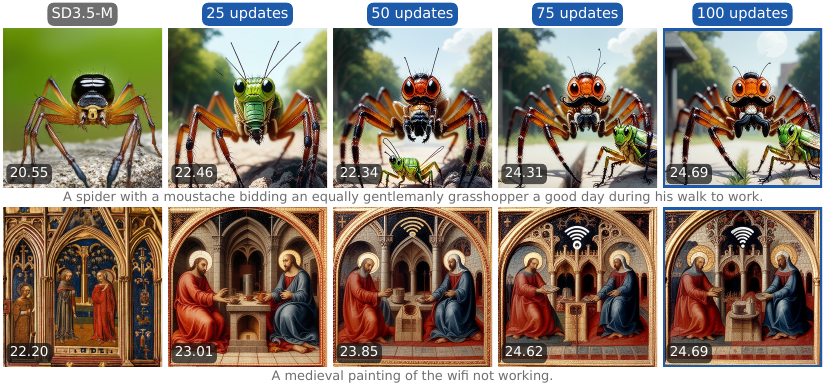}{\linewidth}{2.58in}{Two prompts along training}
\caption{SD3.5-M and \mend{} after 25, 50, 75 and 100 updates, with the same initial noise along each row and PickScore on each tile. Top: the grasshopper appears by update 50 and the moustache by update 75. Bottom: the wifi symbol appears at update 50 and is sharp by update 75.}
\label{fig:progress}
\end{figure}

Reward keeps improving while base distance barely changes late in training: between updates \res{early.update} and \hpUpdatesO{}, DrawBench PickScore rises from \res{cfg45c025.pick} to \mgPick{}, while base distance moves only from \res{cfg45c025.dist} to \mgDist{} (Table~\ref{tab:main}, Figure~\ref{fig:dynamics}(c)). Figure~\ref{fig:progress} follows two seeds through this run.

\subsection{Other rewards and backbones}\label{sec:rewards}
\begin{figure}[t]
\begin{minipage}[t]{0.61\linewidth}
\vspace{0pt}
\centering
\mendfig{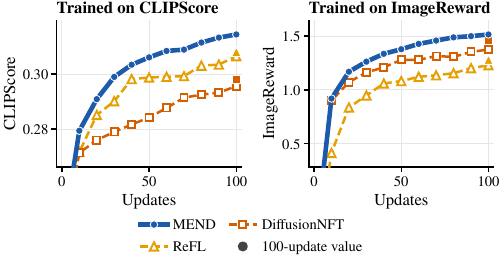}{\linewidth}{1.75in}{Trained reward against updates for CLIPScore and ImageReward}
\caption{Trained reward against updates under the equal-budget protocol of \citet{zhou2026opsd}, one \mend{} run per reward, against ReFL and DiffusionNFT under the same protocol; markers give their 100-update values.}
\label{fig:rewards}
\end{minipage}\hfill
\begin{minipage}[t]{0.37\linewidth}
\vspace{0pt}
\centering
\mendfig{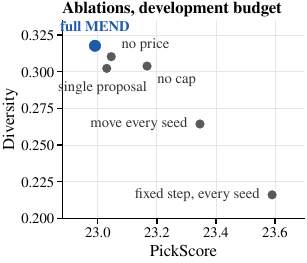}{\linewidth}{1.75in}{One-factor arms at the development budget}
\caption{\textbf{Restricting moves trades reward for diversity.} One-factor arms at the development budget on the development split, with the fixed-step control.}
\label{fig:ablation}
\end{minipage}
\end{figure}

\begin{table}[t]
\centering
\caption{\mend{} surpasses ReFL and DiffusionNFT on every training reward. Each \mend{} run uses SD3.5-M, \hpUpdatesO{} updates and the protocol of \citet{zhou2026opsd}; the base uses the same evaluation setting. The last two columns give ReFL and DiffusionNFT under this protocol on the row's training reward. Bold: \mend{}'s trained metric; underline: second best among the three methods on that metric.\looseness=-1}
\label{tab:transfer}
\vspace{2pt}
\small
\setlength{\tabcolsep}{4pt}
\resizebox{\linewidth}{!}{%
\begin{tabular}{@{}lcccccc@{}}
\toprule
\headrow Training reward & PickScore & HPSv2.1 & ImageReward & CLIPScore & ReFL & DiffusionNFT \\
\midrule
\textcolor{gray}{none (SD3.5-M)} & \textcolor{gray}{\res{abbase.pick}} & \textcolor{gray}{0.207} & \textcolor{gray}{$-0.52$} & \textcolor{gray}{0.239} & & \\
\oursrow PickScore & \best{\mPick} & \mHPSt & \mIR & \mClip & \second{\res{refl.pick}} & \res{nftequal.pick} \\
\oursrow ImageReward & \res{rwimagereward.pick} & \res{rwimagereward.hps2} & \best{\res{rwimagereward.ir}} & \res{rwimagereward.clip} & \res{refl.ir} & \second{\res{nftequal.ir}} \\
\oursrow HPSv2.1 & \res{rwhpsv2.pick} & \best{\res{rwhpsv2.hps2}} & \res{rwhpsv2.ir} & \res{rwhpsv2.clip} & \second{\res{refl.hps2}} & \res{nftequal.hps2} \\
\oursrow CLIPScore & \res{rwclipscore.pick} & \res{rwclipscore.hps2} & \res{rwclipscore.ir} & \best{\res{rwclipscore.clip}} & \second{\res{refl.clip}} & \res{nftequal.clip} \\
\bottomrule
\end{tabular}}
\end{table}
\mend{} surpasses ReFL and DiffusionNFT on every training reward we test. Table~\ref{tab:transfer}, Figure~\ref{fig:rewards} and Figure~\ref{fig:hero}(c) compare separate runs trained on PickScore, ImageReward, HPSv2.1 or CLIPScore under the equal-budget protocol. Each run is above both baselines on its training reward at update \hpUpdatesO{} and at every evaluated update. All four runs finish above the base on every held-out evaluator; three show held-out declines before update \hpUpdatesO{} (Appendix~\ref{app:rewards}). CLIPScore and ImageReward training yield the lowest held-out PickScore (\res{rwclipscore.pick} and \res{rwimagereward.pick}); Appendix~\ref{app:overopt} examines the declines under longer training.

\mend{} is not tied to one backbone: it applies to any flow model with a differentiable reward, and it improves every backbone we test at the same update budget. On SD3-M, PickScore rises from \res{sd3base.pick} to \res{sd3mend.pick} (Table~\ref{tab:sd3m}), above Linear-DPO \citep{li2026lineardpo} at \res{sd3dpo.pick}, while held-out HPSv2.1 rises from \res{sd3base.hps2} to \res{sd3mend.hps2}. On Z-Image-Turbo \citep{cai2025zimage}, a distilled model sampled in nine steps at 1024 pixels, PickScore training raises that score from \res{zbase.pick} to \res{zpick.pick} and held-out HPSv2.1 from \res{zbase.hps2} to \res{zpick.hps2}. HPSv2.1 training reaches \res{zhps.hps2} with held-out PickScore \res{zhps.pick}. Appendix~\ref{app:rewards} gives full results (Tables~\ref{tab:sd3m} and~\ref{tab:zimage}) and training trajectories.

\subsection{Ablations}\label{sec:ablations}
The cap and the price decide which samples receive a move, and together they preserve diversity: against moving every seed below the cap, the full rule keeps 0.037 more diversity for 0.35 PickScore (Table~\ref{tab:ablation}). We vary one component at the development budget (\hpPromptsE{} prompts $\times$ \hpImagesE{} images per update, \hpUpdatesE{} updates, the earlier optimizer setting). Table~\ref{tab:ablation} in Appendix~\ref{app:ablations} evaluates the checkpoints on full DrawBench; Figure~\ref{fig:ablation} uses the development split, where the fixed-step control is available. \mend{} accepts target moves for 34\% of seeds per round, reaching PickScore \res{abref.pick} at diversity \res{abref.div}. Removing the cap or the price raises this share to 45\% and 78\%, respectively, and lowers diversity to \res{abnocap.div} and \res{abnoprice.div}. Assigning a move to every seed below the cap (81\%) yields the highest full-set PickScore, \res{abnvtwo.pick}, and lowest diversity, \res{abnvtwo.div}. At the full per-update budget, varying the initial price, using one candidate or removing the behavior EMA changes update-50 PickScore by at most 0.06; five candidates lower it by 0.25 (Appendix~\ref{app:sensitivity}).

\begin{figure}[!t]
\centering
\mendfig{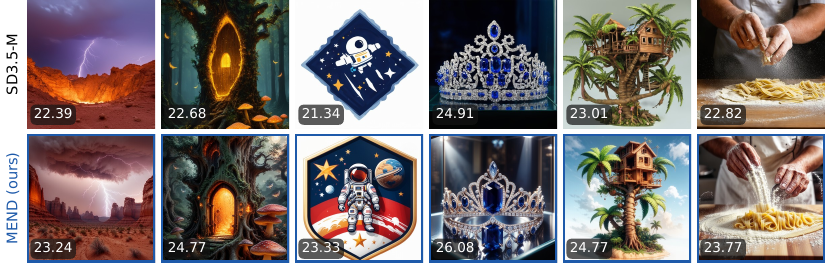}{\linewidth}{1.75in}{Additional matched SD3.5-M and \mend{} pairs}
\caption{More SD3.5-M and \mend{} results.}
\label{fig:qual}
\end{figure}

\subsection{Qualitative results}\label{sec:qual}
\begin{wrapfigure}[26]{r}{0.5\linewidth}
\vspace{-10pt}
\centering
\mendfig{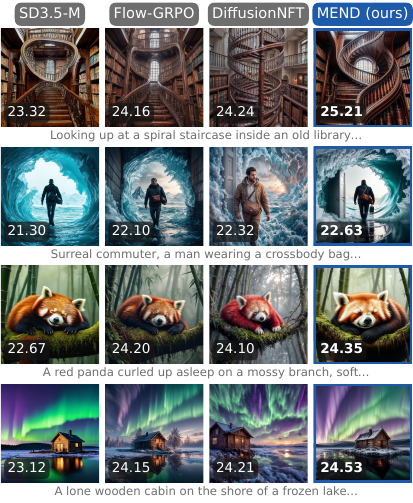}{\linewidth}{3.7in}{More matched prompts across SD3.5-M, Flow-GRPO, DiffusionNFT, and \mend{}}
\caption{\textbf{Matched comparisons with Flow-GRPO and DiffusionNFT.} Same prompt and seed per row; PickScore on each tile, best in bold.}
\label{fig:cmp2}
\end{wrapfigure}

In matched comparisons, \mend{} corrects prompt errors of the base while keeping its composition. Figures~\ref{fig:hero}, \ref{fig:cmp}, \ref{fig:fidelity}, \ref{fig:progress}, \ref{fig:qual} and~\ref{fig:cmp2} use matched prompts and initial noise, with PickScore on each tile. Figures~\ref{fig:cmp} and~\ref{fig:cmp2} compare \mend{} with Flow-GRPO and DiffusionNFT; Figure~\ref{fig:qual} shows base-to-\mend{} pairs. Many \mend{} examples have higher saturation, reflecting a reward preference examined in Appendix~\ref{app:overopt}. Appendix~\ref{app:samples} gives full prompts, additional seeds, an uncurated sheet (Figure~\ref{fig:hires-uncurated-1}) and training progressions.

\section{Related work}\label{sec:related}

DDPO and DPOK optimize rewards with policy gradients, while Diffusion-DPO fits offline preferences \citep{black2023ddpo,fan2023dpok,wallace2024dpo}. Flow-GRPO and DanceGRPO extend group-relative policy gradients to flow models with stochastic sampling and a KL penalty \citep{liu2025flowgrpo,xue2025dancegrpo}. DiffusionNFT, AWM and FlowAWR use reward- or advantage-weighted regression on model-generated samples \citep{zheng2025nft,xue2025awm,fu2026flowawr}. These methods adjust the contribution of existing samples; \mend{} instead constructs new local targets and verifies each one (Appendix~\ref{app:related}).

ReFL, DRaFT and AlignProp backpropagate differentiable rewards through sampling \citep{xu2023imagereward,clark2024draft,prabhudesai2023alignprop}; RAFT and related methods select among existing samples \citep{dong2023raft,lee2025dav}. Concurrently, \citet{zhou2026opsd} regress clean predictions toward bounded reward-gradient targets using group weights, while Self-OPD distills the model's own stochastic branches \citep{zhang2026selfopd}. \mend{} accepts a proposal only when its capped reward gain over the unchanged sample exceeds a quadratic displacement penalty. This per-target check differs from distribution-level Wasserstein regularization \citep{fan2025orwcfm,hwang2026wgf}.

\section{Conclusion}\label{sec:conclusion}
\mend{} post-trains flow models by proximal velocity matching: a sample's target moves only when its capped reward gain exceeds a quadratic displacement price, and the model regresses onto the resulting velocity targets. This one per-sample rule replaces KL penalties, frozen reference models and advantage weights. In \hpUpdatesO{} updates, \mend{} outperforms Flow-GRPO on every evaluator except aesthetic score at the same distance to base-model images, and it surpasses ReFL and DiffusionNFT at equal budget on four rewards. \mend{} is backbone-agnostic and easy to adopt: it improves SD3.5-M, SD3-M and the distilled Z-Image-Turbo at the same update budget. At \res{train.gpuh} GPU-hours for the SD3.5-M run, reward post-training becomes cheap enough to repeat for each new reward or backbone.

\textbf{Limitations.} \mend{} requires a differentiable reward. Its guarantees hold for the targets of each round. Longer training can over-optimize a reward, and some single-reward runs lose held-out score within the \hpUpdatesO{}-update budget (Appendices~\ref{app:overopt} and~\ref{app:toy-rounds}).

\setlength{\bibsep}{6pt plus 2pt minus 2pt}
\bibliography{refs}
\bibliographystyle{mend_arxiv}

\newpage
\appendix
\input{appendix/v2_appendix}
\end{document}